\documentclass[letterpaper]{article} % DO NOT CHANGE THIS

\usepackage{amsmath}  
\usepackage{aaai25}  % DO NOT CHANGE THIS
\usepackage{times}  % DO NOT CHANGE THIS
\usepackage{helvet}  % DO NOT CHANGE THIS
\usepackage{courier}  % DO NOT CHANGE THIS
\usepackage[hyphens]{url}  % DO NOT CHANGE THIS
\usepackage{graphicx} % DO NOT CHANGE THIS
\usepackage{natbib}  % DO NOT CHANGE THIS AND DO NOT ADD ANY OPTIONS TO IT
\usepackage{caption} % DO NOT CHANGE THIS AND DO NOT ADD ANY OPTIONS TO IT
\usepackage{multirow}     % For merged rows (used in the "Model" column)
\usepackage{array}        % For better table formatting
\usepackage{booktabs}  
\usepackage{amsmath}      % For math symbols
\usepackage{siunitx}      % For better number formatting in tables

\usepackage{booktabs}     % For better table formatting
\usepackage{multirow}     % For merged cells
\usepackage{threeparttable}  % For table notes
\usepackage{siunitx}      % For number alignment

\usepackage{caption}      % For better caption formatting
\usepackage{graphicx}
\usepackage{fancyhdr}

\usepackage{algorithm}
\usepackage{algorithmic}

\usepackage{amsmath}
\usepackage{newfloat}
\usepackage{listings}
\DeclareCaptionStyle{ruled}{labelfont=normalfont,labelsep=colon,strut=off} % DO NOT CHANGE THIS
\floatstyle{ruled}
\newfloat{listing}{tb}{lst}{}
\floatname{listing}{Listing}
\newcommand{\uiuc}[1]{{#1\textsuperscript{\includegraphics[scale=0.25]{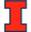}}}}
\newcommand{\lgicon}[1]{{#1\textsuperscript{\includegraphics[scale=0.003]{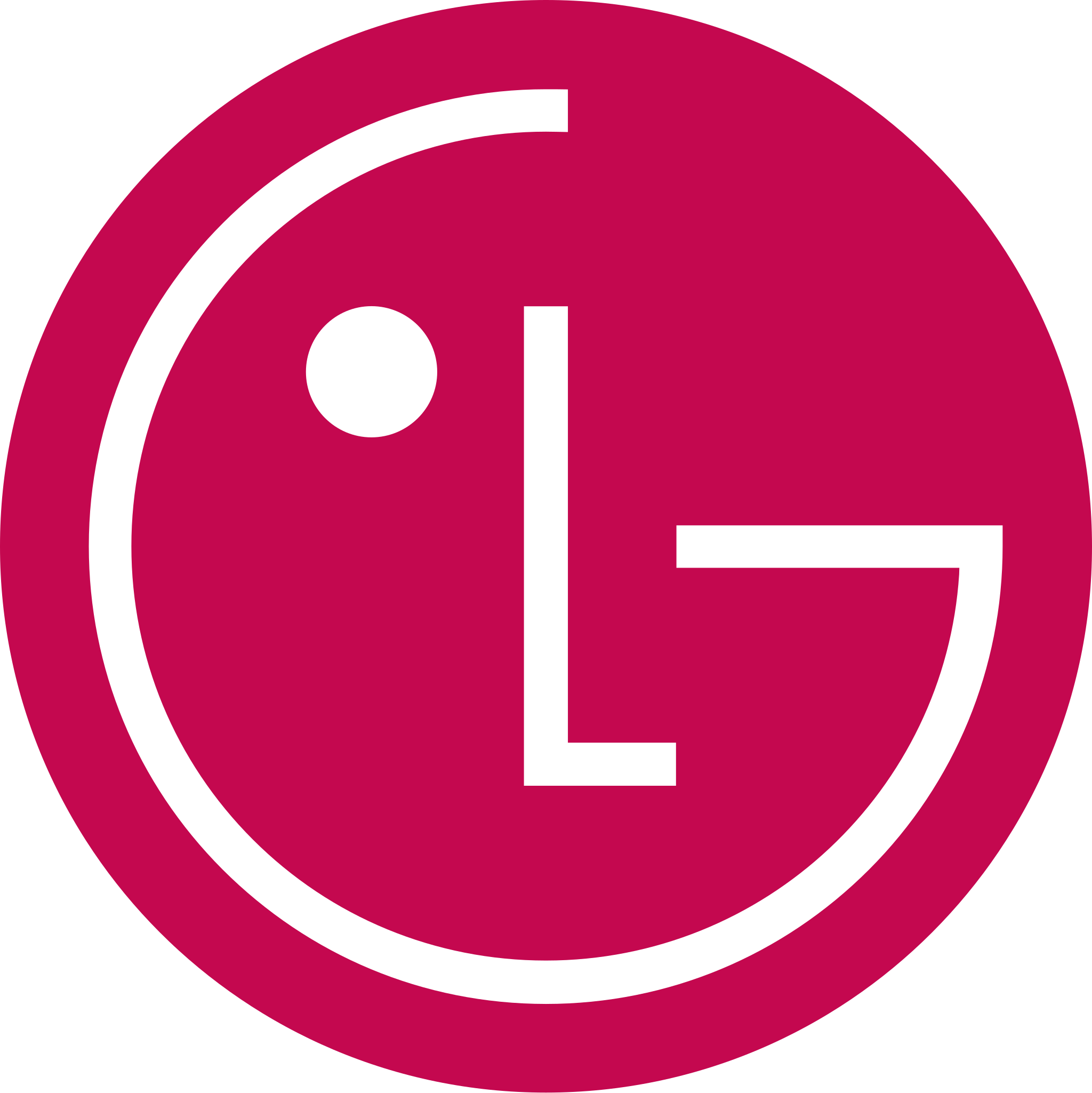}}}}

\title{Clear Preferences Leave Traces: Reference Model-Guided Sampling for Preference Learning}
\author{
    \uiuc{Nirav Diwan\footnote{ \footnotesize \textit{corresponding author - ndiwan2@illinois. Nirav worked on this project during the summer internship at LG AI Research \\}}},
    \lgicon{Tolga Ergen},
    \lgicon{Dongsub Shim},
    \lgicon{Honglak Lee}
}
\affiliations{
    \uiuc{}University of Illinois Urbana-Champaign
    \lgicon{}LG AI Research\\
}

\usepackage{bibentry}

\begin{document}
% \linenumbers

\maketitle
\begin{abstract}
Direct Preference Optimization (DPO) has emerged as a de-facto approach for aligning language models with human preferences. 
Recent work has shown DPO's effectiveness relies on training data quality. In particular, clear quality differences between preferred and rejected responses enhance learning performance. 
Current methods for identifying and obtaining such high-quality samples demand additional resources or external models. 
We discover that reference model probability space naturally detects high-quality training samples. Using this insight, we present a sampling strategy that achieves consistent improvements ($+0.1$ to $+0.4$) on MT-Bench while using less than half ($30$-$50\%$) of the training data.
We observe substantial improvements ($+0.4$ to $+0.98$) for technical tasks (coding, math, and reasoning) across multiple models and hyperparameter settings.

\end{abstract}

\section{Introduction}
Preference learning aims to align Large Language Models (LLMs) with human preferences. 
It is applied after pre-training and supervised fine-tuning (SFT) to teach models to generate responses that better align with human expectations while preserving knowledge acquired in earlier training stages. 
This approach has demonstrated practical impact in improving user experience~\cite{bai2024aligning}, implementing safety filters~\cite{liu2024enhancing, huang2024one}, and moderating content~\cite{ma2023adapting}.

\vspace{2mm}

Direct Preference Optimization (DPO)~\cite{rafailov2024direct} is a a supervised off-policy method that has recently emerged as a leading approach to preference learning. 
Unlike on-policy methods, DPO directly optimizes the policy using paired preference data, where each pair contains an aligned (or \textit{preferred}) and misaligned (or \textit{rejected}) response. 
By training the model to assign higher probabilities to preferred responses, DPO effectively shapes the model's output distribution without requiring a separate reward model. 
This simplicity, combined with strong empirical results, has made DPO increasingly popular for language model alignment.

\vspace{2mm}

Models learn better when there is a clear distinction between preferred and rejected responses, while noisy or ambiguous preferences hinder learning~\cite{ivison2024unpacking}.
This distinction is quantified using \textit{preference clarity}, measured using ground truth scalar values that indicate each response's alignment or instruction-following ability. 
Prior work has explored various approaches to address this challenge: modifying alignment  objectives~\cite{gao2024impact}, filtering training data~\cite{morimura2024filtered}, and improving preference data collection~\cite{morimura2024filtered}.

\begin{figure}[t] % Use [t] for top placement within a single column
    \centering
    \includegraphics[width=\columnwidth]{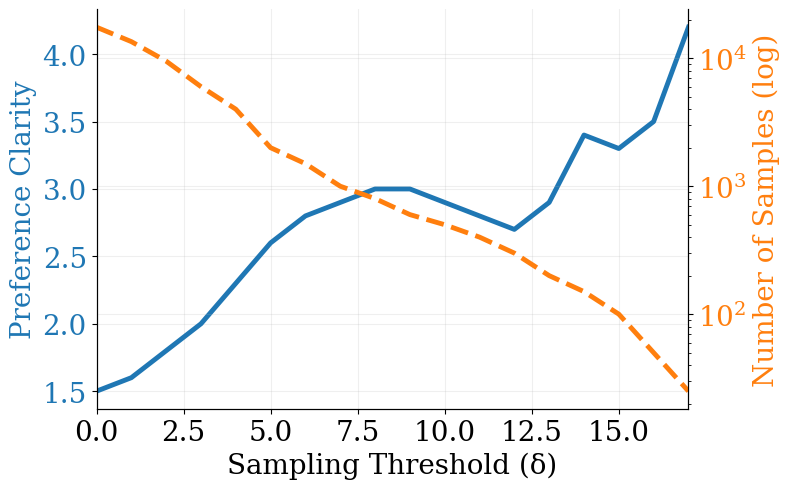} % Adjust width to fit one column
    \caption{Relationship between sampling threshold ($\delta$) and preference clarity for Ultrafeedback using the fine-tuned LLAMA-3 8B as the reference model.
    The solid blue line is the preference clarity between preferred and rejected responses calculated using the difference of Ultrafeedback's preference scores.
    The dashed orange line (log scale) shows the available training pairs at each threshold.
    \textbf{Preference pairs at a higher sampling threshold show clearer preferences, indicating that the reference model can identify high-quality preference pairs even when it incorrectly attributes the correct response.}
    }
    \label{fig:clarity_threshold}
\end{figure}

\vspace{2mm}

% \begin{figure*}[h]
%     \centering
%     \includegraphics[width=\textwidth]{images/main_image.pdf}
%     \caption{Reference model-based sampling leverages probability differences to identify clear preference signals. (a) Examples of clear vs. ambiguous preferences. (b) Clear preferences naturally separate in reference model probability space. (c) This guides sampling toward stronger training signals.}
%     \label{fig:example}
% \end{figure*}

Creating high-quality preference pairs requires extensive annotation resources. 
Current datasets like Ultrafeedback~\cite{cui2023ultrafeedback} use GPT-4 for evaluation, but scaling this approach to new alignment datasets increases costs and creates dependencies on external models. 
The better way to ensure high-quality annotation is to perform manual annotation. 
However, this demands considerable effort — requiring multiple rounds of review, quality checks, and careful measurement of annotator agreement. 
These resource limitations drive the need for methods that can identify high-quality preference pairs without expensive labeling processes.

\vspace{2mm}

We find that the probability space of the reference model serves as a natural detector of preference clarity. 
When there is strong gap between the probabilities of the two responses — regardless of which it favors — the pair represents a clearer preference signal in the alignment data. 
This reveals an intriguing property: \textit{the reference model can identify high quality preference pairs, even when it may not know which response is better}.
Using this insight, we propose a reference model-guided sampling method that achieves better performance than the full dataset of sampling method.

% We validate our findings by experimenting on multiple hyperparameter settings and model architectures.
% %
% We observe that the improvement is robust across different hyperparameter settings and model architectures, suggesting we're detecting fundamental signals rather than exploiting model-specific patterns.
% %
% Specifically, sampling around $30\%$ - $50\%$ of the original training data we achieve higher performance on MT-Bench ($0.1 - 0.4$), particularly on technical tasks (coding, reasoning, math) ($0.4 - 0.98$).
% %
% We ablate our hyperparameters and suggest mechanisms for adopting this strategy for.

\vspace{2mm}

We validate our findings through extensive experimentation across model architectures and hyperparameter configurations. 
The improvements remain consistent across these variations, suggesting our method captures fundamental preference signals rather than exploiting model-specific patterns. 
Using only $30$-$50\%$ of the original training data, we achieve higher MT-Bench performance ($+0.1$ to $+0.4$), with particularly strong gains on technical tasks ($+0.4$ to $+0.98$ for coding, reasoning, and math). 
Through ablation studies, we identify key factors for effective adoption of this sampling strategy.

\section{Methodology}
\subsection{Background}

\paragraph{Direct Preference Optimization}(DPO) ~\cite{rafailov2024direct} is a supervised off-policy method used to optimize models based on user preferences without relying on a separate reward model. Instead, DPO directly aligns the policy with preference data by leveraging a ranking-based approach.

\vspace{2mm}

The objective of DPO is to use the reference policy, typically a SFT model, to guide the optimization process. 
The preference optimization is defined using pairs of responses, where one response is preferred over the other. By focusing on these response pairs, the goal is to minimize the DPO loss function defined as:

\begin{align*}
\mathcal{L}_{\text{DPO}}(\pi_\theta; \pi_{\text{ref}}) = -{E}_{(x, y_w, y_l) \sim \mathcal{D}} 
\Bigg[ & \log \sigma \Bigg( \beta \log \frac{\pi_\theta(y_w | x)}{\pi_{\text{ref}}(y_w | x)} \\
& - \beta \log \frac{\pi_\theta(y_l | x)}{\pi_{\text{ref}}(y_l | x)} \Bigg) \Bigg]
\end{align*}

where $\pi_{\theta}$ is the policy model, $\pi_{\text{ref}}$ is the reference model, $\sigma$ is the sigmoid function, and $(x, y_w, y_l)$, $\beta$ is the hyperparameter controlling the deviation from the base reference policy, are preference pairs comprising a prompt $x$,  a preferred response $y_w$, and a rejected response $y_l$, drawn from the dataset $\mathcal{D}$.

\paragraph{Preference Clarity} It quantifies the degree of quality difference between two responses in a preference pair. Given responses $r_1$ and $r_2$, we calculate preference clarity using ground truth quality scores $s_1$ and $s_2$ assigned to each response. These scores can be obtained either through human annotation or using LLM-as-a-judge frameworks with models like GPT-4. Formally, for a preference pair $(r_1, r_2)$, the preference clarity is measured as:
\begin{equation}
   clarity(r_1, r_2) = |s_1 - s_2|
\end{equation}
where $s_1, s_2$ are the ground truth quality scores.

\subsection{Reference-model based sampling}

\paragraph{Method.} Our sampling strategy leverages normalized reference policy probabilities to identify clear preference signals.
For input $x$ with responses $y_w$ (preferred) and $y_l$ (rejected), we compute the difference between length-normalized reference probabilities. We sample pairs where this difference exceeds threshold $\delta$:

\vspace{1mm}

\begin{equation}
\label{eqn:main}
    \left|\frac{\log(\pi_{\text{ref}}(y_w|x))}{|y_w|} - \frac{\log(\pi_{\text{ref}}(y_l|x))}{|y_l|}\right| \geq \delta
\end{equation}

\vspace{1mm}

    where $\pi_{\text{ref}}(y|x)$ is the reference probability and $|y|$ is the sequence length. This normalization enables comparison between responses of different lengths, and focuses training on pairs with larger probability gaps.
This simple approach  prioritizes training examples with clearer preference signals while downweighting pairs where reference model estimates that the responses are similar.

\paragraph{Reference Model as Quality Detector.} Our analysis reveals that the sampling strategy effectively identifies high-quality preference pairs. 
Using LLAMA-3-8B as the reference model, we compute probability differences between preferred and rejected responses in the Ultrafeedback dataset (Figure~\ref{fig:clarity_threshold}).
Samples selected at higher thresholds ($\delta$) in Equation~\ref{eqn:main} consistently demonstrate greater preference clarity. 
This suggests reference models naturally detect strong training examples without requiring additional annotation.

% Recent analysis of DPO's learning dynamics reveals a notable insight: the model primarily amplifies existing probability differences rather than creating new ones~\cite{chen2024preference}. 
% %
% When the reference model assigns similar probabilities to both responses in a pair, DPO typically struggles to learn a meaningful preference. 
% %
% Conversely, initial probability gaps tend to widen during training, suggesting that these gaps indicate inherent preference clarity. 
% %
% This observation led us to investigate whether reference model probability could directly identify strong training signals, bypassing the need for additional clarity measures or objective modifications. 
%
% Figure X shows the distribution of normalized probability differences between chosen and rejected responses, with the average labeled preference scores. 
% %
% When the absolute difference ($\delta$) is small (0-2), the average GPT4 score difference is also small (1.5-2.0), indicating ambiguous preferences. 
% %
% We observe consistently higher GPT4 score differences (2.6-6.5) as $\delta$ increases indicating clearer preference signals. 
% %
% This relationship, while not perfectly correlated ($r=0.3$), enables us to identify higher-quality training pairs without requiring additional labeling or a separate reward model.

% Place this where you want the table
\begin{table*}[t]
\centering
\begin{threeparttable}
% \caption{Impact of Reference Model Probability Gap Thresholds on Model Performance}
\begin{tabular*}{\textwidth}{@{\extracolsep{\fill}}lccccc}
\toprule
& \multicolumn{2}{c}{\textbf{Dataset}} & \multicolumn{2}{c}{\textbf{MT-Bench Performance}} \\
\cmidrule(lr){2-3} \cmidrule(lr){4-5}
\textbf{Model} & $\delta$ & Percentage & Score & $\Delta$ vs Full Dataset\tnote{\textdagger} \\
\midrule
\multirow{5}{*}{Mistral 7B} 
& SFT & -- & $6.33$ & $-0.63$ \\
& \hspace{1em} + $\delta \geq 0$ & $100\%$ & $6.96$ & -- \\
& \hspace{1em} + $\delta \geq 0.5$ & $70\%$ & \textbf{$7.13$} & \textbf{$+0.17$} \\
& \hspace{1em} + $\delta \geq 1$ & $48\%$ & $6.97$ & $+0.01$ \\
& \hspace{1em} + $\delta \geq 2$ & $21\%$ & $6.99$ & $+0.03$ \\
\midrule
\multirow{4}{*}{LLAMA-3-8B} 
& SFT & -- & $6.55$ & $-0.95$ \\
& \hspace{1em} + $\delta \geq 0$ & $100\%$ & $7.40$ & -- \\
& \hspace{1em} + $\delta \geq 1$ & $57\%$ & $7.77$ & $+0.37$ \\
& \hspace{1em} + $\delta \geq 2$ & $31\%$ & $7.8$ & $+0.40$ \\
\midrule
\multirow{5}{*}{LLAMA-3.1-8B} 
& SFT & -- & $6.56$ & $-1.21$ \\
& \hspace{1em} 

+ $\delta \geq 0$ & $100\%$ & $7.74$ & -- \\
& \hspace{1em} + $\delta \geq 1$ & $57\%$ & $7.88$ & $+0.14$ \\
& \hspace{1em} + $\delta \geq 2$ & $48\%$ & $7.85$ & $+0.11$ \\
\bottomrule
\end{tabular*}
\begin{tablenotes}
\small
\item[\textdagger] $\Delta$ shows performance difference compared to using full dataset ($\delta \geq 0$). 
\caption{All the models achieve better performance ($0.1 - 0.4$) on MT-Bench with the sampled data using our reference-model sampling approach compaed to the full dataset ($\delta \geq 0$) and the SFT model.}
\label{tab:performance}
\end{tablenotes}
\end{threeparttable}
\end{table*}

\section{Experiments}

We use a simple evaluation framework for our sampling method. 
First, we use an SFT model as our reference model. 
We compute the probabilities for all responses in our preference pairs. 
We then create different versions of our training dataset by sampling based on probability gaps. 
Each version uses a different threshold $\delta$. 
We align models using DPO on both the full dataset and our sampled versions. 
Finally, we evaluate and compare their performances on standard benchmarks. 
This setup directly tests whether selecting clearer preference pairs improves model alignment.

\paragraph{Dataset.} We use the Ultrafeedback dataset~\cite{cui2023ultrafeedback} for all our experiments. 
The dataset provides a binarized preference version containing $64$k samples, where each response pair is accompanied by preference scores. 
These scores are generated using GPT4 in an LLM-as-a-judge framework~\cite{zheng2023judging}, which evaluates responses across multiple technical aspects using scalar ratings. 
This scoring approach has demonstrated high agreement with human annotators on technical criteria. 
We leverage the difference between these preference scores as a ground truth measure of preference clarity.

\paragraph{Models and Hyperparameters} We perform experiments using three different models - Mistral 7B~\cite{jiang2023mistral}, LLAMA-3-8B~\cite{dubey2024llama} and LLAMA-3.1-8B~\cite{dubey2024llama}. 
For LLAMA-3, we use the SFT checkpoint provided by SimPO ~\cite{meng2024simpo} after training on Ultrachat~\cite{ding2023enhancing}
For Mistral and LLAMA-3.1, we use the Ultrachat dataset to perform SFT on the models.
For table \ref{tab:performance}, we perform all our experiments with $\beta = 0.01$ for DPO.
We experiment with different versions of the reference thresholds ($\delta = 0, 0.5, 1, 2$).

\paragraph{Evaluation} We measure the performance of the model on MT-Bench~\cite{zheng2023judging}, a multi-turn alignment dataset that grades the answers of the policy model using GPT4 by assigning a scalar score to each response.
MT-Bench categorizes their questions on 8 categories - writing, roleplay, extraction, reasoning, math, coding, knowledge I (STEM), and knowledge II (humanities/social science).

\section{Results and Discussion}

\subsection{Performance}
We present the results in table \ref{tab:performance}.
The policy models trained using our proposed reference model-based sampling consistently outperform those trained on the full dataset.
Notably, the improvements range from moderate to large ($+0.11$ to $+0.40$), with most notable gains achieved using reduced training data.
For LLAMA3, we observe large improvements ($+0.40$) while using less than a third of the original dataset. 
Similarly, LLAMA 3.1 shows moderate gains ($+0.14$) using less than half the data. 
The pattern differs slightly for Mistral 7B, where the largest performance improvement occurs with $70\%$ of the original dataset and a smaller sampling threshold.

\subsection{Hyperparameters}
The performance of our approach depends on both the sampling threshold ($\delta$) and preference optimization parameters.

\paragraph{Sampling Threshold ($\delta$):} No single sampling threshold $\delta$ works optimally across all models. 
While higher thresholds identify clearer preference pairs, they also reduce the training data size, which can limit performance gains.
For instance, Mistral 7B achieves optimal performance at $\delta = 0.5$, with larger thresholds showing no slight improvement over the full dataset. 
We note that our discrete sampling thresholds ($\delta \in \{0, 0.5, 1, 2\}$) may not exhaustively cover the optimal values for each model, suggesting potential for further optimization.
However, we typically notice the largest increase in benchmark performance when 50-70 \% of the original dataset is retained using the reference-based sampling.

\vspace{2mm}

\paragraph{Preference Optimization ($\beta$):} In contrast, we believe a fixed $\beta = 0.01$ performs well regardless of the sampling threshold.
We notice small increases ($+0.1$) in MT-Bench for LLAMA-3.1-8B when we train it using large $\beta = 0.1$.
However, we do not observe an increase in performance for LLAMA-3-8B.
Instead, the performance remains the same or slightly decreases ($-0.1$) for sampling thresholds.
One possible explanation may be that a larger $\beta$ imposes a stronger penalty on model deviations, potentially limiting beneficial adaptations, while a smaller $\beta$ permits more flexibility, which may enhance alignment with human preferences. 

\begin{figure}[h]
    \centering
    \includegraphics[width=\columnwidth]{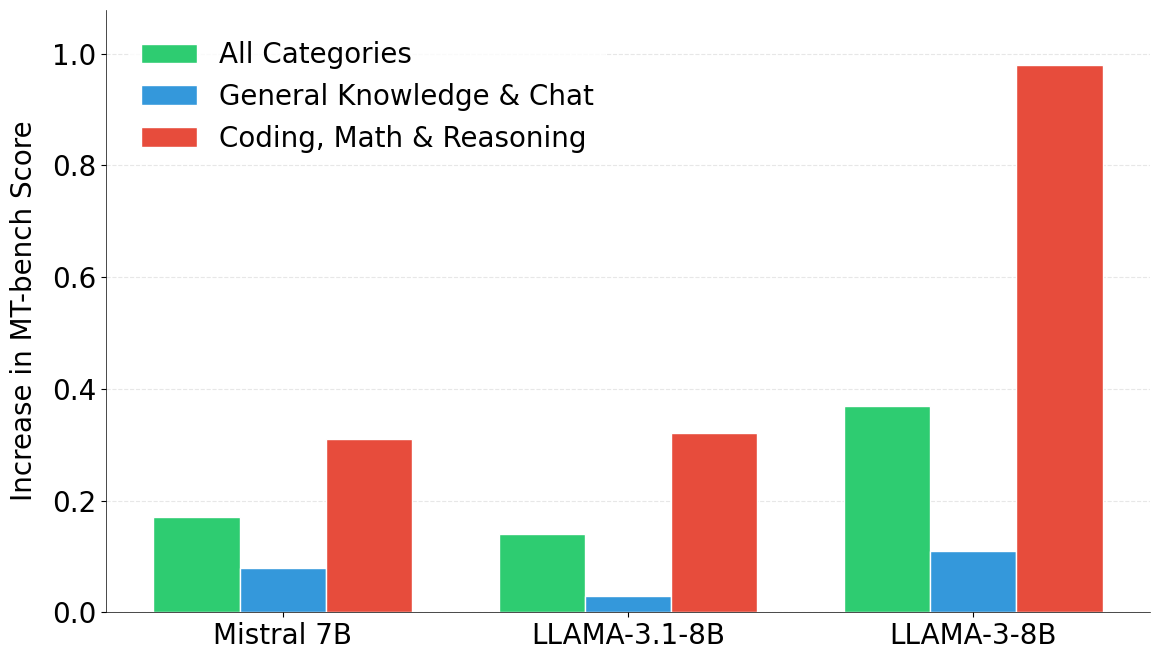}
    \caption{Performance improvements across different task categories for the best version of sampling approach , measured by increase in MT-bench scores. Technical tasks (Coding, Math \& Reasoning) show substantially larger gains compared to general tasks.}
    \label{fig:mt_bench_incrase}
\end{figure}

\subsection{Aspect-wise increases}
We also analyze task-specific performance improvements on MT-Bench. 
Technical tasks (coding, math \& reasoning) show large gains compared to general knowledge and chat tasks (writing, roleplay, extraction, knowledge). 
LLAMA3-8B demonstrates the most dramatic improvements, with MT-Bench scores for technical tasks increasing by nearly $1.0$ points. 
While Mistral and LLAMA-3.1 show more modest gains, their improvements on technical tasks remain substantial ($+0.4$). 
The improvements in general knowledge and chat tasks, while consistent across models, are less pronounced ($+0.2$ to $+0.4$). 
This disparity might be attributed to benchmark saturation -- models trained on the full dataset already achieve high scores on non-technical tasks, leaving limited room for improvement. 
In contrast, technical tasks present more headroom for measurable gains, suggesting our sampling method is effective at identifying and leveraging strong preference signals for technical tasks.

\section{Limitations \& Future Work}
There are several promising directions to build upon our work. First, testing our approach on LLMs of varying sizes and architectures would help verify the conditions under which this property holds. 
Experiments with models trained on different SFT datasets could further establish the generality of our findings. 
While MT-Bench serves as a standard benchmark for measuring alignment and instruction-following capabilities, it has limitations in consistency, reliability and biases~\cite{zheng2023judging}. 
Alternative benchmarks like Arena-hard~\cite{li2024crowdsourced} could provide additional validation, though computational costs currently restrict such evaluation. 
It would also be interesting to verify if this property holds for other preference learning mechanisms.
From a theoretical perspective, understanding why alignment techniques learn better from samples with large probability gaps in the reference model space could provide insights into preference learning mechanisms.
\section{Related Work}

\paragraph{Preference Learning Methods.}
Preference alignment has emerged as a crucial step in developing LLMs for production environments. Reinforcement Learning from Human Feedback (RLHF)~\cite{ouyang2022training, christiano2017deep} pioneered this approach, using on-policy learning to align models with human preferences. Recent work has shifted toward off-policy methods, with Direct Preference Optimization (DPO)~\cite{rafailov2024direct} gaining widespread adoption due to its simplicity and effectiveness. 
Several variants of DPO have been proposed, introducing additional regularization terms~\cite{liu2023statistical, pal2024smaug}, new hyperparameters~\cite{meng2024simpo}, or modified training objectives~\cite{ethayarajh2024kto}. 

\paragraph{Quality-Focused Approaches.}
The challenge of obtaining high-quality preference pairs has been addressed through three main approaches.
The first focuses on improving data collection, using either human experts or LLMs to curate training samples~\cite{hu2024towards, huang2023learning, jiang2024survey}. 
The second develops more robust training mechanisms through modified objectives~\cite{wu2024alpha, chowdhury2024provably}. 
These approaches have seen limited adoption compared to standard DPO. 
The third approach involves collecting large datasets and filtering out noisy samples~\cite{morimura2024filtered, kim2024aligning}.  
While data collection improvements require manual oversight, and robust training methods have shown limited practical success, the filtering approach offers a promising direction -- especially for collecting large scale data and then narrowing down the high quality samples, preventing re-iteration and expensive use of resources.

\paragraph{Data Quality in Alignment.}
Recent work has highlighted how preference data quality impacts alignment success~\cite{ivison2024unpacking}. 
Models demonstrate enhanced learning from clearly distinguished preference pairs, motivating research into methods for identifying and generating high-quality training data. 
For instance, FilterDPO~\cite{morimura2024filtered} proposes an on-policy approach that selects training samples by comparing policy-generated responses with preferred responses based on reward differences. 
While effective, such methods require additional computation or reward modeling. 
Our work presents a complementary approach that identifies strong preference pairs using only reference model probabilities, eliminating the need for additional resources or inference steps.

% \section{Acknowledgments}

\bigskip
\noindent 
\bibliography{aaai25}

\end{document}